\documentclass[a4paper,twoside]{article}

\usepackage{epsfig}
\usepackage{subcaption}
\usepackage{calc}
\usepackage{amssymb}
\usepackage{amstext}
\usepackage{amsmath}
\usepackage{amsthm}
\usepackage{multicol}
\usepackage{pslatex}
\usepackage{apalike}
\usepackage{algorithm2e}
\usepackage[bottom]{footmisc}

\usepackage{booktabs}
\usepackage{xcolor}
\usepackage{tikz}
\usetikzlibrary{arrows}
\usetikzlibrary{arrows.meta}

\usepackage[pagebackref=true,breaklinks=true,colorlinks,bookmarks=false]{hyperref}

\usepackage{SCITEPRESS}     

\definecolor{aliceblue}{rgb}{0.94, 0.97, 1.0}
\definecolor{honeydew}{rgb}{0.94, 1.0, 0.94}
\definecolor{lightyellow}{rgb}{1.0, 1.0, 0.88}
\definecolor{platinum}{rgb}{0.9, 0.89, 0.89}

\DeclareMathOperator*{\argmax}{arg\, max}

\begin{document}

\title{Multi-Scale Foreground-Background Confidence for~Out-of-Distribution~Segmentation}

\author{\authorname{Samuel Marschall\sup{1} and Kira Maag\sup{2}}
\affiliation{\sup{1}Technical University of Berlin, Germany}
\affiliation{\sup{2}Heinrich-Heine-University Düsseldorf, Germany}
\email{s.marschall@tu-berlin.de, kira.maag@hhu.de}
}

\keywords{Deep Learning, Computer Vision, Out-of-Distribution Segmentation, Foreground-Background Segmentation.}

\abstract{
Deep neural networks have shown outstanding performance in computer vision tasks such as semantic segmentation and have defined the state-of-the-art. However, these segmentation models are trained on a closed and predefined set of semantic classes, which leads to significant prediction failures in open-world scenarios on unknown objects. As this behavior prevents the application in safety-critical applications such as automated driving, the detection and segmentation of these objects from outside their predefined semantic space (out-of-distribution (OOD) objects) is of the utmost importance. In this work, we present a multi-scale OOD segmentation method that exploits the confidence information of a foreground-background segmentation model. While semantic segmentation models are trained on specific classes, this restriction does not apply to foreground-background methods making them suitable for OOD segmentation. We consider the per pixel confidence score of the model prediction which is close to 1 for a pixel in a foreground object. By aggregating these confidence values for different sized patches, objects of various sizes can be identified in a single image. Our experiments show improved performance of our method in OOD segmentation compared to comparable baselines in the SegmentMeIfYouCan benchmark.
}

\onecolumn \maketitle \normalsize \setcounter{footnote}{0} \vfill

%
%
%
\section{\uppercase{Introduction}}
\begin{figure}[t]
    \center
    \includegraphics[trim=0 0 0 0,clip,width=0.46\textwidth]{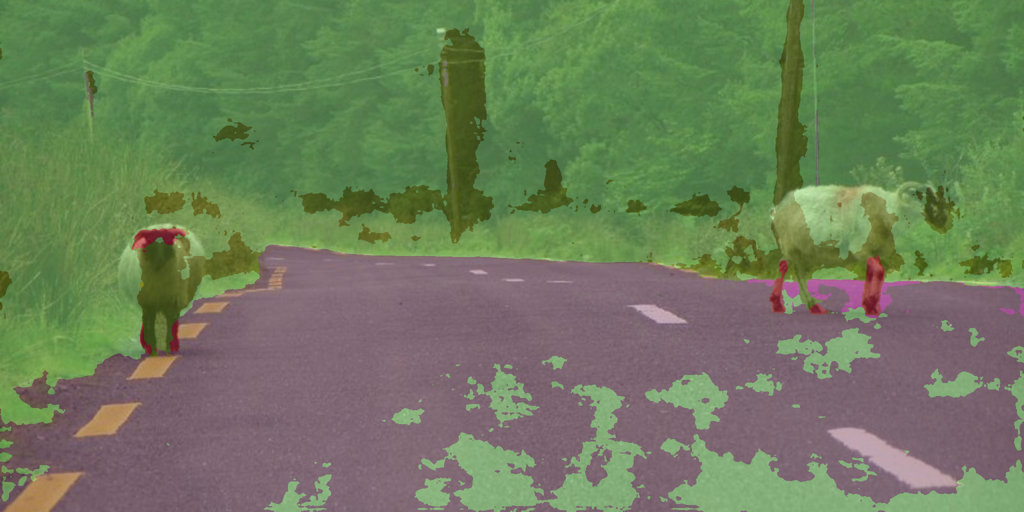}
    \includegraphics[trim=0 0 0 0,clip,width=0.46\textwidth]{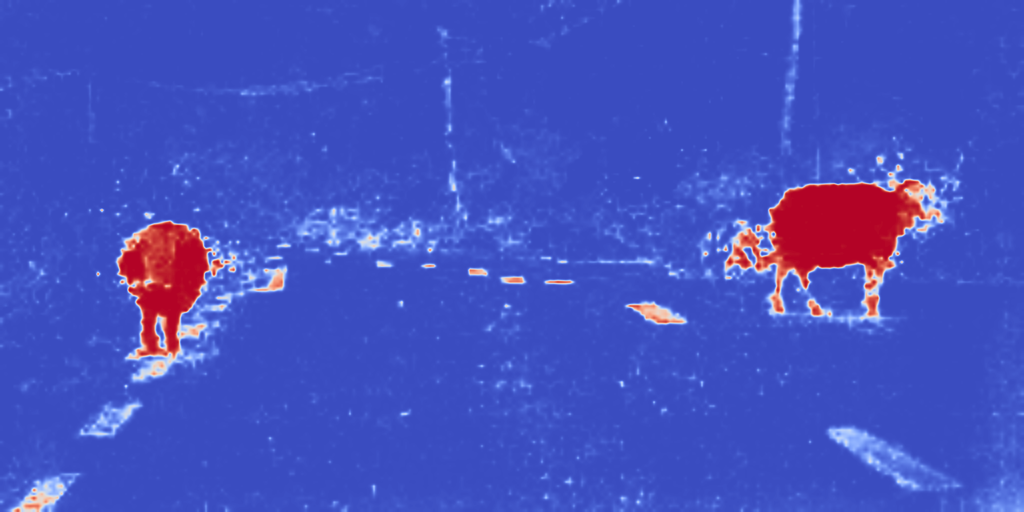}
    \caption{\emph{Top}: Semantic segmentation predicted by a DNN. \emph{Bottom}: Confidence heatmap obtained by our method.}
    \label{fig:seg_ood}
\end{figure}

Deep neural networks (DNNs) have demonstrated outstanding performance in computer vision tasks like image classification \cite{Wortsman2022}, object detection \cite{Wang2023}, instance segmentation \cite{Yan2023} and semantic segmentation \cite{Xu2023}. These computer vision tasks are also frequently used in safety-critical areas such as medical diagnosis and automated driving. In the latter case, information about the environment, i.e., an understanding of the scene, is of highest importance and can be provided by e.g. semantic segmentation (pixels of an input image are decomposed into segments which are assigned to a fixed and predefined set of semantic classes). 
Currently, the leading method \cite{Hummer2023} for semantic segmentation on the Cityscapes test dataset \cite{Cordts2016}, which represents a street scenario of dense urban traffic in various German cities, achieves a strong mean intersection over union score of 86.4\%. 
However, the performance of DNNs degrades rapidly in open-world scenarios on unseen objects for which the network has not been trained. An example of two sheep crossing a road is shown in Figure~\ref{fig:seg_ood} (top).
These objects from outside the network's semantic space are called \emph{out-of-distribution} (OOD) objects. On the one hand, these can really be new object classes, such as animals, or, on the other hand, objects that belong to a known class but appear significantly different from other objects of the same class seen during training. 
Independent of the object type, it is crucial to detect and segment them to protect the network from incorrect and dangerous predictions. The computer vision task of identification and segmentation of these objects is referred to as \emph{OOD segmentation}~\cite{Chan2021_1,Maag2023_dataset}.

A broad area of OOD segmentation methods are uncertainty-based techniques that do not use retraining or OOD data. 
A well-known approach to estimate model uncertainty is Bayesian modeling \cite{Mackay1992}. 
Monte Carlo Dropout (MC Dropout, \cite{Gal2016}) is computationally feasible for computer vision tasks, thus serving as an approximation for Bayesian models, and has already been applied to semantic segmentation \cite{Lee2020}. A similar idea is to use ensemble learning to capture uncertainty, i.e., averaging predictions over multiple sets of parameters \cite{Lakshminarayanan2017}. 
In addition to these sampling strategies, there are also uncertainty estimation methods that are based only on the output of the DNN, for example maximum softmax probability \cite{Hendrycks2016}. In \cite{Maag2024_grads}, the pixel-wise gradient norms were introduced. The magnitude features of gradients at inference time provide information about the uncertainty propagated in the corresponding forward pass. 
A benchmark for uncertainty estimation in the real-world task of semantic segmentation for urban driving is presented in \cite{Blum2019} where pixel-wise uncertainty estimates are evaluated towards the detection of anomalous objects in front of the ego-car. 

In this work, we introduce a multi-scale OOD segmentation method that exploits the confidence information of a foreground-background segmentation model on patches of different sizes and aggregates them into a single OOD score map. 
The terms confidence and uncertainty are directly linked, as confidence describes how strongly the model believes that its prediction is correct, while uncertainty describes the degree of doubt the model has about its prediction. Both concepts are based on the predicted output probabilities of the network. 
Note, our approach also does not require any additional training or auxiliary data. 
An overview of our method is given in Figure~\ref{fig:method}. 
The latest foreground-background models \cite{Simeoni2023,Wang2023c} are trained in a self-supervised way on datasets like ImageNet \cite{Deng2009} to generate a binary mask indicating foreground or background. 
Datasets like ImageNet are also commonly used to train vision backbones.
In comparison to supervised semantic segmentation models using a closed set of predefined classes, this independence from class prediction makes it reasonable to apply these models to the detection of unknown objects. 
Since these models focus on images in which one or more objects are present and these differ from the background, the prediction of foreground objects in real street scenes is more complex and the classification into foreground and background is not necessarily unambiguous \cite{Maag2023_domain}. For this reason, we propose a multi-scale approach to detect OOD objects of different sizes and also show different approaches to aggregate these patches of various sizes. 
Moreover, we do not use the binary output of the model but the per pixel confidence score of the model prediction which is close to 1 for a pixel in a foreground object. 
By aggregating these confidence scores across the different patches, different sized objects in a single image can be identified. 
Furthermore, it has already been shown in other works that uncertainty information extracted from the softmax output of a DNN for semantic segmentation is appropriate for segmenting OOD objects, especially if the input images resemble the training images and only contain additional unknown objects. 
Thus, we combine our foreground-background confidence heatmap with the pixel-wise softmax uncertainty of a semantic segmentation network. 

Our contributions can be summarized as follows:
\begin{itemize}
    \item We introduce a new OOD segmentation method based on confidence information of a foreground-background model.
    \item We describe various approaches for the multi-scale procedure, i.e., how the image patches of different sizes can be constructed and combined.
    \item We show how and in which cases our confidence approach can be supported by uncertainty information from a semantic segmentation network. 
    \item We demonstrate the effectiveness of our method for OOD segmentation on different OOD datasets outperforming a variety of comparable (uncertainty-based) methods.
\end{itemize}
%
%
%
\section{\uppercase{Related work}}\label{sec:rel_work}
Various uncertainty methods have already been tested for OOD segmentation, including maximum softmax probability \cite{Hendrycks2016} and sampling-based methods such as MC Dropout \cite{Mukhoti2020} and deep ensembles \cite{Lakshminarayanan2017}. In \cite{Maag2024_grads}, the pixel-wise gradient norms (PGN) were introduced which provide information about the uncertainty propagated in the corresponding forward pass. 
The methods Mahalanobis \cite{Lee2018} and ODIN \cite{Liang2018} enhance the separation of softmax score in- and out-of-distributions by performing adversarial attacks on the input images. 

Other OOD segmentation methods do not consider the output of the DNN, instead focusing on the feature space. 
Density estimation of the in-distribution feature representations is conducted in \cite{Galesso2023} using a nearest-neighbor approach. 
An online data condensation algorithm is presented in \cite{Vojir2024} which extracts a pixel/patch feature representation and builds a two-dimensional projection space to find the optimal and calibrated in- and out-of-distribution decision strategy. 
In \cite{Sodano2024}, two decoders are used to push features of pixels belonging to the same class together, one decoder produces a Gaussian model for each known category and the other performs binary anomaly segmentation.
The method described in \cite{Ackermann2023} accesses neither the feature space nor the pure output, but the raw mask prediction of a mask-based semantic segmentation network. These networks also learn to assign certain masks to anomalies but such masks are discarded by default when generating semantic predictions.

Another line of research relies on the exploitation of OOD data for training, which is disjoint from the original training data \cite{Biase2021,Blum2019_1,Chan2021,Gao2023,Grcic2022,Grcic2023,Liu2023,Rai2023,Tian2022}. 
However, the additional data does not have to be real-world data, rather it can be created synthetically. 
This synthetic negatives are for example used to reduce energy in negative pixels \cite{Nayal2023}. In \cite{Delic2024}, an ensemble of in-distribution uncertainty and the posterior of the negative class formulate a novel outlier score. 
This type of research also includes works that use the normalizing flow to generate the negative data \cite{Blum2019_1,Grcic2021,Gudovskiy2023}. 

Alternative methods for OOD segmentation use complex auxiliary models. In \cite{Besnier2021}, an image is perturbed by a local adversarial attack and the observer network is trained to predict network's errors. 
Discrepancy networks are used in \cite{Lis2019,Lis2020} to compare the original image and the resynthesized one highlighting the unexpected objects. 
To recognize and reconstruct road, a reconstruction module is trained in \cite{Vojir2021,Vojir2023}, as a poor reconstruction is due to areas that are outside of the training distribution. 
In \cite{Zhang2024}, a pixel decoder, a transformer decoder, a base teacher network and multilayer perceptrons are trained together for OOD detection.

Specialized training approaches for OOD segmentation rely on various types of retraining with additional data, i.e., OOD data or synthetically generated data, which often requires generative models. 
Our method does not need retraining, OOD data or complex auxiliary models. 
Our approach is more similar to classical approaches quantifying uncertainty, since we only use in-distribution data and rely on the output of the network. In particular, our method is comparable to sampling methods (MC Dropout and deep ensembles) due to our multi-scale approach, even though we do not use average predictions over multiple sets of parameters. 
In addition, we compare with Mahalanobis and ODIN, whose goal is to calibrate the softmax score, even if the computation of adversarial samples requires a full backward pass.
%
%
%
\section{\uppercase{Method description}}\label{sec:method}
\begin{figure*}
    \centering
    \resizebox{0.9\linewidth}{!}{\input{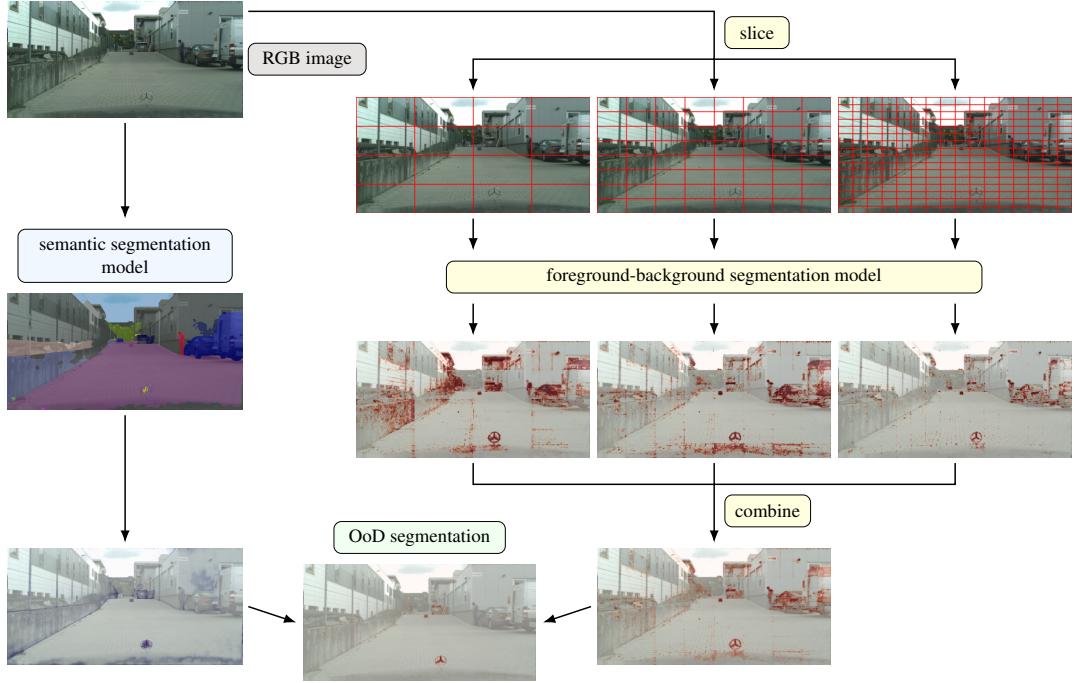}}
    \caption{Schematic illustration of our multi-scale OOD segmentation method. On the one hand, the input image is divided into different sized slices, inferred by the foreground-background model and the confidence heatmaps are aggregated into a single output map. On the other hand, the image is fed into a semantic segmentation network, which outputs an uncertainty heatmap, which is then combined with the confidence map of the foreground-background model to obtain the final OOD segmentation.}
    \label{fig:method}
\end{figure*}
In this section, we describe our OOD detection method, which consists of two branches, i.e., the multi-scale foreground-background segmentation and the semantic segmentation. An overview of our approach is shown in Figure~\ref{fig:method}. 
%
%
\paragraph{Multi-scale Foreground-Background Segmentation.} 
In the first branch, the input images are divided into patches and fed into the foreground-background model. Let $x\in\mathbb{R}^{H\times W\times 3}$ denote the input image in RGB-format with height $H$ and width $W$. We reshape this image into a sequence of patches $\mathcal{X} = \{ x_n \in \mathbb{R}^{P_H\times P_W\times 3} \}_{n=1,\ldots,N}$ where $P_H\times P_W$ describes the patch size and $N = \frac{H}{P_H} \cdot \frac{W}{P_W}$ the number of patches. 
Here, we assume that the patches do not overlap and that the image is completely covered by the patches, i.e., they do not extend over the boundary of the image. Otherwise we can just perform a suitable resize operation first. 
Each single image patch $x_n$ is then fed into a foreground-background segmentation model which outputs $\theta_n (x_n) \in[0,1]^{P_H\times P_W}$ providing the predicted probability for each pixel to belong to a foreground object using the sigmoid function. 
Since the individual patches do not overlap, we can simply reassemble all these individual masks to obtain a mask $\theta (x) \in[0,1]^{H\times W}$ that covers the complete original image. Note, assuming that the patches do not overlap, their sizes do not have to be the identical, but can vary within a single image.

Since the OOD objects in an image can differ in size, we use patches of different sizes. This means that we completely divide the input image into equally sized patches of size $P_H^i \times P_W^i$ and then divide the image again into equally sized patches, but with a different scale ($P_H^{i+1} \times P_W^{i+1}$). 
In total, we obtain a set of predicted masks from the foreground-background model $\{ \theta^i (x) \in [0,1]^{H\times W} \}_{i=1,\ldots,d}$ where $d$ describes the number of different scales. 
In Figure~\ref{fig:method}, for example, we have used three different scales ($d=3$), with $N$ (number of patches per image) having the values 16, 64 and 256, respectively.
The predicted masks are combined by 
\begin{equation}
    \hat{\theta} (x) = \sum_{i=1}^d \alpha^i \theta^i (x) \in [0,1]^{H\times W} \enspace, 
\end{equation}
where $\alpha^i\in[0,1]$, $i=1,\dots,d$, with $\sum_{i=1}^{d} \alpha^i = 1$. 
This aggregated multi-scale prediction provides pixel-wise information about the confidence of each pixel belonging to a foreground object, i.e., values close to 0 or 1 indicate high confidence that there is no object or there is an object at that position, whereas values close to 0.5 indicate high uncertainty. Here, foreground corresponds with OOD object, as OOD objects are located on the street and are therefore clearly in the foreground.
%
%
\paragraph{Semantic Segmentation.} 
In the other branch, we consider a semantic segmentation network. Each pixel $z$ of an input image $x$ gets assigned a label $y$ from a prescribed label space $C = \{y_{1}, \ldots, y_{c} \}$. Given learned weights $w$, the DNN provides for the $z$-th pixel a probability distribution $f(x;w)_{z} \in [0,1]^{|C|}$ consisting of the probabilities for each class $y \in C$ denoted by $p(y|x)_z \in [0,1]$. 
Note, the predicted class is then computed by $\hat y_{z}^x =\argmax_{y\in\mathcal{C}} p(y|x)_z$.

The uncertainty in the semantic segmentation prediction is quantified by the commonly used entropy, which is defined by
\begin{equation}
    E(x)_{z} =- \frac{1}{\log(|C|)} \sum_{y\in \mathcal{C}}p(y|x)_{z} \cdot \log p(y|x)_{z} \in [0,1] \enspace,
\end{equation}
whereby the fraction only serves for normalization. 
Entropy values close to 1 indicate high uncertainty, as all classes are equally distributed, which may indicate an unknown class, i.e., an OOD pixel. 
Since OOD objects are mainly located on the street, another uncertainty measure that can indicate these objects is the predicted probability for the class ``road''. High values indicate OOD objects, thus, we use
\begin{equation}
    R(x)_{z} =1- p(y=\text{``road''}|x)_{z} \in [0,1]
\end{equation}
as another uncertainty heatmap. 

The idea is that this uncertainty information from the semantic segmentation network supports the foreground-background model in OOD segmentation, especially in cases where these images are similar to the training data. 
The final confidence map for an image $x$ is computed by 
\begin{equation}\label{mult}
    \hat{\theta} (x) * D(x) \in [0,1]^{H\times W}, \ D \in \{ E,R \} \enspace,
\end{equation}
using the component-wise multiplication. 
Note, the focus of our method is on the multi-scale confidence of the foreground-background model, so this second branch is optional.
%
%
%
\section{\uppercase{Experiments}}\label{sec:exp}
First, we present the experimental setting and then study our method in terms of its OOD segmentation capability. 
%
%
\subsection{Experimental Setting}\label{sec:exp_setting}
%
%
\paragraph{Segmentation Models.} 
For the foreground-background segmentation, we consider two recent models, FOUND \cite{Simeoni2023} and CutLER \cite{Wang2023c}. Both methods use the vision transformer DINO \cite{Caron2021} as basis and are self-supervised trained on the ImageNet dataset \cite{Deng2009}, which provides image data for training large-scale object recognition. 
Note, the ImageNet dataset is also frequently used for the backbone training of semantic segmentation networks. 
For semantic segmentation, we use the state-of-the-art DeepLabv3+ network \cite{Chen2018} with ResNet-101 backbone \cite{He2016}. This DNN is trained on the Cityscapes dataset \cite{Cordts2016} achieving a mean intersection over union (mIoU) of 80.21\%. The Cityscapes dataset consists of dense urban traffic from various German cities. 
%
%
\paragraph{OOD Datasets.} 
To evaluate the OOD segmentation performance of our method, we consider the three datasets from the SegmentMeIfYouCan benchmark\footnote{\url{http://segmentmeifyoucan.com/}}. 
The LostAndFound dataset \cite{Pinggera2016} includes 1,203 validation images, with annotations marking the road surface and the OOD objects, specifically small obstacles on German roads positioned in front of the ego-vehicle. A refined version, LostAndFound test-NoKnown, is available in \cite{Chan2021_1}. 
The RoadObstacle21 dataset \cite{Chan2021_1} consists of 412 test images and is similar to the LostAndFound dataset, as all obstacles are positioned on the street. However, it offers greater diversity in both, the OOD objects and situational contexts. 
Meanwhile, the RoadAnomaly21 dataset \cite{Chan2021_1}, containing 100 test images, presents various unique objects (anomalies) that can appear anywhere within the image. 
%
%
\paragraph{Evaluation Metrics.}
To assess the OOD segmentation performance, we follow the evaluation protocol of the official SegmentMeIfYouCan benchmark.
For the evaluation on pixel-level, the threshold-independent area under the precision-recall curve (AuPRC) is used which measures the separability between OOD and not OOD. In addition, the false positive rate at 95\% true positive rate (FPR$_{95}$) serves as safety-critical metric of how many false positive errors must be made in order to achieve the desired rate of true positive predictions. 
For the segment-level evaluation, the set of metrics includes an adjusted version of the mIoU (sIoU) to assess segmentation accuracy at a specific threshold, the positive predictive value (PPV or precision) as binary instance-wise accuracy and the $F_1$-score. 
These segment-wise metrics are averaged across thresholds from 0.25 to 0.75 in steps of 0.05, yielding $\overline{\text{sIoU}}$, $\overline{\text{PPV}}$ and $\overline{F_1}$. 
%
%
\subsection{Numerical Results}\label{sec:exp_results}
\begin{table}[t]
\caption{OOD segmentation results for the FOUND model applied to LostAndFound with different number of patches $N$ as well as combinations of these confidence maps.}
\centering
\scalebox{0.70}{
\begin{tabular}{l cc ccc}
\toprule 
 & \multicolumn{5}{c}{LostAndFound test-NoKnown} \\
\cmidrule(r){2-6} 
& AuPRC $\uparrow$ & FPR$_{95}$ $\downarrow$ & $\overline{\text{sIoU}}$ $\uparrow$ & $\overline{\text{PPV}}$ $\uparrow$ & $\overline{F_1}$ $\uparrow$ \\
\midrule
1    & 44.5 & 44.4 &  7.7 & 18.1 &  6.1 \\
16   & 47.4 & 23.8 & 19.4 & 23.0 & 15.6 \\
64   & 45.6 & 18.9 & 27.8 & 21.2 & 16.4 \\
256  & 38.4 & 21.4 & 26.7 & 17.4 & 10.1 \\
1024 & 24.6 & 40.6 & 26.5 & 16.0 &  6.5 \\
\midrule
16+64            & 52.0 & 15.0 & 24.0 & 24.3 & 18.0 \\
16+64+256        & 54.0 & 11.8 & 29.9 & 22.3 & 18.8 \\
16+64+256+1024   & \textbf{56.5} & \textbf{10.6} & \textbf{32.2} & 22.5 & \textbf{19.5} \\
1+16+64+256      & 54.0 & 12.7 & 24.9 & 23.1 & 17.4 \\
1+16+64+256+1024 & 56.4 & 11.7 & 26.0 & \textbf{25.2} & 18.7 \\
\bottomrule
\end{tabular} }
\label{tab:ood_laf_found}
\end{table}
\begin{table}[t]
\caption{OOD segmentation results for the CutLER model applied to LostAndFound with different number of patches $N$ as well as combinations of these confidence maps.}
\centering
\scalebox{0.70}{
\begin{tabular}{l cc ccc}
\toprule 
 & \multicolumn{5}{c}{LostAndFound test-NoKnown} \\
\cmidrule(r){2-6} 
& AuPRC $\uparrow$ & FPR$_{95}$ $\downarrow$ & $\overline{\text{sIoU}}$ $\uparrow$ & $\overline{\text{PPV}}$ $\uparrow$ & $\overline{F_1}$ $\uparrow$ \\
\midrule
1   & 22.5 & 100 & 6.4 & \textbf{56.5} & 13.0 \\
16  & 18.1 & 100 & 1.7 & 35.2 &  3.1 \\
64  & 22.3 & 100 & 2.9 & 45.8 &  5.1 \\
256 & 18.2 & 100 & 3.0 & 50.2 &  4.6 \\
\midrule
1+64          & 30.9 & \textbf{54.3} & \textbf{18.7} & 42.2 & \textbf{24.8} \\
1+16+64       & 33.8 & 66.2 & 15.5 & 43.4 & 22.8 \\
1+16+64+256   & \textbf{37.1} & 63.6 & 13.7 & 50.0 & 19.9 \\
\bottomrule
\end{tabular} }
\label{tab:ood_laf_cutler}
\end{table}
\begin{table*}[t]
\caption{OOD segmentation results for the multi-scale FOUND model applied to LostAndFound in combination with uncertainty heatmaps obtained by the semantic segmentation network.}
\centering
\scalebox{0.70}{
\begin{tabular}{l cc ccc cc ccc}
\toprule 
\multicolumn{1}{c}{} & \multicolumn{5}{c}{entropy $E$} & \multicolumn{5}{c}{road probability $R$} \\ 
\cmidrule(r){2-6} \cmidrule(r){7-11}
& AuPRC $\uparrow$ & FPR$_{95}$ $\downarrow$ & $\overline{\text{sIoU}}$ $\uparrow$ & $\overline{\text{PPV}}$ $\uparrow$ & $\overline{F_1}$ $\uparrow$ & AuPRC $\uparrow$ & FPR$_{95}$ $\downarrow$ & $\overline{\text{sIoU}}$ $\uparrow$ & $\overline{\text{PPV}}$ $\uparrow$ & $\overline{F_1}$ $\uparrow$\\
\midrule 
16+64            & 75.7 & 4.9 & 36.1 & 41.8 & 35.6     & 63.9 & 6.6 & 27.0 & 35.6 & 25.1 \\
16+64+256        & 77.1 & 4.4 & 39.2 & 41.7 & 37.8     & 65.5 & \textbf{6.3} & 29.0 & 37.0 & 26.9 \\
16+64+256+1024   & \textbf{77.6} & \textbf{4.2} & \textbf{40.6} & 43.5 & \textbf{39.5}     & \textbf{65.9} & \textbf{6.3} & \textbf{29.9} & 37.3 & \textbf{27.6} \\
1+16+64+256      & 76.4 & 4.6 & 36.9 & 43.5 & 37.1     & 64.2 & 6.7 & 27.1 & 36.8 & 25.7 \\
1+16+64+256+1024 & 76.9 & 4.5 & 37.8 & \textbf{44.7} & 38.3     & 64.6 & 6.7 & 27.5 & \textbf{38.0} & 26.5 \\
\bottomrule
\end{tabular} }
\label{tab:ood_laf_found_entro}
\end{table*}
\begin{table*}[t]
\caption{OOD segmentation results for the FOUND model applied to RoadAnomaly21 and RoadObstacle21 with different number of patches $N$ as well as combinations of these confidence maps.}
\centering
\scalebox{0.70}{
\begin{tabular}{l cc ccc cc ccc}
\toprule 
\multicolumn{1}{c}{} & \multicolumn{5}{c}{RoadAnomaly21} & \multicolumn{5}{c}{RoadObstacle21} \\ 
\cmidrule(r){2-6} \cmidrule(r){7-11}
& AuPRC $\uparrow$ & FPR$_{95}$ $\downarrow$ & $\overline{\text{sIoU}}$ $\uparrow$ & $\overline{\text{PPV}}$ $\uparrow$ & $\overline{F_1}$ $\uparrow$ & AuPRC $\uparrow$ & FPR$_{95}$ $\downarrow$ & $\overline{\text{sIoU}}$ $\uparrow$ & $\overline{\text{PPV}}$ $\uparrow$ & $\overline{F_1}$ $\uparrow$\\
\midrule 
1    & \textbf{81.9} & 11.0 & 48.0 & 23.8 & 10.6     & 62.3 & 24.5 & 13.1 & 25.1 & 13.3 \\
4    & 75.5 & 17.1 & 54.1 & 22.2 &  8.5     & 76.5 & 17.8 & 27.6 & \textbf{43.3} & 31.3 \\
16   & 68.4 & 27.0 & 60.3 & 15.3 &  5.7     & 70.8 & 12.9 & 38.4 & 36.4 & 32.6 \\
64   & 60.6 & 42.0 & 59.6 & 10.6 &  3.0     & 56.0 & 64.1 & 39.5 & 24.5 & 20.8 \\
256  & 43.7 & 68.8 & 46.3 &  9.4 &  1.0     & 29.9 & 84.4 & 30.8 & 18.7 &  8.9 \\
\midrule
1+4            & 81.0 &  9.9 & 53.8 & \textbf{26.7} & \textbf{13.6}     & 76.5 & 14.1 & 21.2 & 42.7 & 25.8 \\
1+4+16         & 80.2 &  \textbf{8.8} & 58.8 & 21.2 & 12.3     & 81.4 &  4.4 & 30.9 & 38.3 & 29.8 \\
1+4+16+64      & 78.9 &  8.9 & \textbf{64.1} & 16.1 & 10.1     & \textbf{81.6} &  \textbf{2.0} & 39.7 & 34.0 & \textbf{32.9} \\
16+64+256      & 65.4 & 26.1 & 62.9 & 10.6 &  3.9     & 64.6 & 11.4 & \textbf{45.2} & 25.3 & 22.2 \\
\bottomrule
\end{tabular} }
\label{tab:ood_ao_found}
\end{table*}
%
%
%
\paragraph{Comparison of Foreground-Background Models.} 
First, we compare the two foreground-background models using different scales, i.e., patch sizes. To this end, we feed the image as a whole into the models ($N=1$), as well as with different numbers of patches $N$, whereby the patches per scale have the same size (see Figure~\ref{fig:method}). For FOUND, we consider $N=1,16,64,256,1024$ and for CutLER $N=1,16,64,256$. Note, as CutLER is more computationally expensive, we do not run this model with 1024 patches. 
The results for both models applied to the LostAndFound dataset are shown in the upper section of Tables~\ref{tab:ood_laf_found} and \ref{tab:ood_laf_cutler}, respectively. 
The best results for FOUND are achieved with a number of 16 and 64 patches, and for CutLER when the whole image is fed into the model. This is due to the reason that CutLER focuses on object detection/instance segmentation, while FOUND concentrates on the distinction between foreground and background pixels. This behavior is also reflected in the performance values, as FOUND is convincing in the pixel-wise metrics and CutLER achieves high $\overline{\text{PPV}}$ values. 

In the bottom part of both tables the results for the different combinations of the confidence maps for various scales are given with a uniform weighting, i.e., $\alpha^i= \frac{1}{d}, i=1,\ldots,d$. The idea of these combinations is to match the best performing scales with each other. 
We observe that the multi-scale approach produces significantly improved results for both foreground-background models. Moreover, FOUND clearly outperforms CutLER in three metrics (AuPRC, FPR$_{95}$ and $\overline{\text{sIoU}}$) and achieves similar scores for $\overline{F_1}$. 
For this reason and as CutLER is more expensive computationally, we will only conduct further experiments with FOUND. 

In Appendix~\ref{sec:app_A}, there are further experiments for FOUND applied to LostAndFound, where the combination of the different scales is not uniform, but with different weights $\alpha^i$. Furthermore, we have conducted experiments where the confidence maps of the different scales are not combined, but patches of different sizes are applied in one image. 
Both procedures show no or only minor improvements in individual metrics compared to the results obtained in Table~\ref{tab:ood_laf_found}. 
%
%
\paragraph{Including Uncertainty of Semantic Segmentation.} 
\begin{figure*}[t]
    \center
    \includegraphics[trim=0 0 0 0,clip,width=0.32\textwidth]{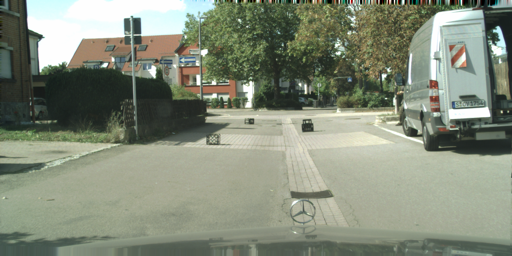}
    \includegraphics[trim=0 0 0 0,clip,width=0.32\textwidth]{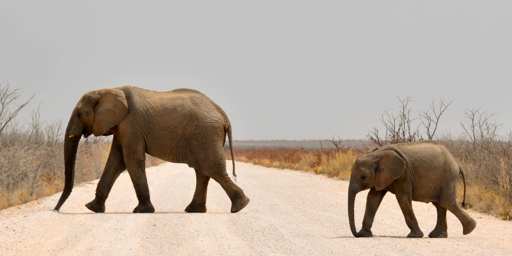}
    \includegraphics[trim=0 15 0 15,clip,width=0.32\textwidth]{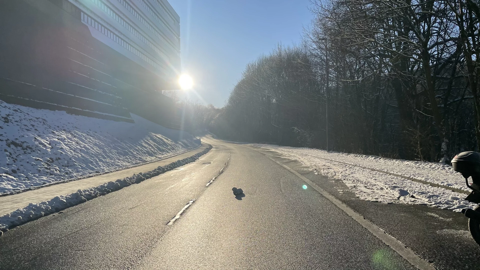} \\
    \vspace{0.5ex}
    \subfloat[][LostAndFound]{\includegraphics[trim=0 0 0 0,clip,width=0.32\textwidth]{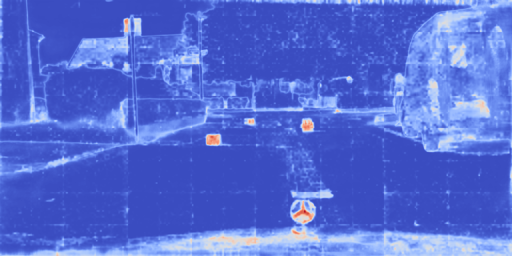} }
    \subfloat[][RoadAnomaly21]{\includegraphics[trim=0 0 0 0,clip,width=0.32\textwidth]{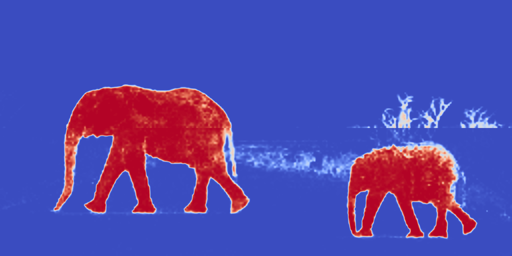} }
    \subfloat[][RoadObstacle21]{\includegraphics[trim=0 15 0 15,clip,width=0.32\textwidth]{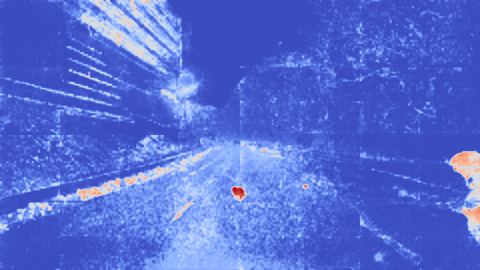} }
    \vspace{1ex}
    \caption{\emph{Top:} RGB images of the LostAndFound, RoadAnomaly21 and RoadObstacle21 dataset. \emph{Bottom:} The corresponding OOD segmentation heatmaps obtained by our method.}
    \label{fig:ood_hm}
\end{figure*}
In this paragraph, we combine the confidence maps of the multi-scale foreground-background model approach with the uncertainty heatmaps obtained by the semantic segmentation model. The results for the multi-scale FOUND model applied to the LostAndFound dataset in combination with the two uncertainty heatmaps, i.e., entropy and the road probability, are given in Table~\ref{tab:ood_laf_found_entro}. 
The additional uncertainty information improves performance even further (in comparison to Table~\ref{tab:ood_laf_found}), independent of the uncertainty heatmap. This might be caused by the fact that the LostAndFound images depict similar street scenes as the in-distribution dataset Cityscapes. 
The highest results are again achieved with a uniform combination of the confidence maps of $N=16,64,256,1024$. 
When comparing the two uncertainty heatmaps, the entropy outperforms the road probability in all metrics. 

%
%
\paragraph{Results for RoadAnomaly21 and RoadObstacle21.} 
The results for the RoadAnomaly21 and the RoadObstacle21 dataset for different number of patches $N$ as well as combinations of these confidence maps are shown in Table~\ref{tab:ood_ao_found}. Since in both datasets, especially in RoadAnomaly21 (see Figure~\ref{fig:ood_hm} (b)), there are also larger OOD in the images, we also consider $N=4$, i.e., divide the inputs into only 4 slices. 
We observe again that the combination of different scales performs better than slicing just one image. 
In Appendix~\ref{sec:app_A}, we also show the results in combination with the entropy as an uncertainty heatmap. For both datasets, the entropy cannot enhance the OOD segmentation performance.
The RoadAnomaly21 dataset differs greatly from Cityscapes, as large objects are in the foreground and the background does not resemble the dense urban traffic in Germany. The RoadObstacle21 dataset contains street scenes but under different conditions (e.g. snow or highway). 
In Figure~\ref{fig:ood_hm}, we show one example per dataset for our OOD segmentation heatmaps derived from the best performing combination, i.e., the uniform combination of the confidence maps of $N=1,4$ for RoadAnomaly21, $N=1,4,16,64$ for RoadObstacle21 and $N=16,64,256,1024$ multiplied with the entropy $E$ for LostAndFound.
%
%
\paragraph{Comparison with Baselines.} 
Here, we compare our method with the comparable baselines from the SegmentMeIfYouCan benchmark, i.e., the uncertainty-based approaches (Maximum Softmax, MC Dropout, Ensemble and PGN) as well as the ones using adversarial attacks (ODIN and Mahalanobis). For our approach, we use the combinations mentioned above, which we have also applied in Figure~\ref{fig:ood_hm}. 
The OOD segmentation results on the LostAndFound dataset are given in Table~\ref{tab:ood_laf} and on the RoadAnomaly21 as well as the RoadObstacle21 dataset in Table~\ref{tab:ood_ao}. 
For LostAndFound, we achieve comparatively high results, similar to the gradient-based PGN method. 
As previously mentioned, the LostAndFound dataset is similar to Cityscapes only with OOD objects, which is why the uncertainty methods perform well. This behavior has also been shown in our experiments where the entropy heatmaps enhance our foreground-background predictions. 
For RoadAnomaly21 and RoadObstacle21, we significantly exceed the baselines, e.g. we obtain AuPRC values of about 81\% for both datasets, which means an increase of more than 44.3 percentage points.
Also worth mentioning are the low FPR$_{95}$ values of 9.9\% and 2.0\%, respectively. This shows that our method is more robust than the uncertainty-based methods when the environment changes.
\begin{table}[t]
\caption{OOD segmentation benchmark results for the LostAndFound dataset.}
\centering
\scalebox{0.70}{
\begin{tabular}{l cc ccc}
\toprule 
 & \multicolumn{5}{c}{LostAndFound test-NoKnown} \\
\cmidrule(r){2-6} 
& AuPRC $\uparrow$ & FPR$_{95}$ $\downarrow$ & $\overline{\text{sIoU}}$ $\uparrow$ & $\overline{\text{PPV}}$ $\uparrow$ & $\overline{F_1}$ $\uparrow$ \\
\midrule
Maximum Softmax    & 30.1 & 33.2 & 14.2 & \textbf{62.2} & 10.3 \\
MC Dropout         & 36.8 & 35.6 & 17.4 & 34.7 & 13.0 \\
Ensemble           & 2.9 & 82.0 & 6.7 & 7.6 & 2.7 \\
PGN                & \underline{69.3} & \underline{9.8} & \textbf{50.0} & 44.8 & \textbf{45.4} \\ 
\midrule
ODIN               & 52.9 & 30.0 & 39.8 & \underline{49.3} & 34.5 \\
Mahalanobis        & 55.0 & 12.9 & 33.8 & 31.7 & 22.1 \\
\midrule
ours               & \textbf{77.6} & \textbf{4.2} & \underline{40.6} & 43.5 & \underline{39.5} \\ 
\bottomrule
\end{tabular} }
\label{tab:ood_laf}
\end{table}
\begin{table*}[t]
\caption{OOD segmentation benchmark results for the RoadAnomaly21 and the RoadObstacle21 dataset.}
\centering
\scalebox{0.70}{
\begin{tabular}{l cc ccc cc ccc}
\toprule 
\multicolumn{1}{c}{} & \multicolumn{5}{c}{RoadAnomaly21} & \multicolumn{5}{c}{RoadObstacle21} \\ 
\cmidrule(r){2-6} \cmidrule(r){7-11}
& AuPRC $\uparrow$ & FPR$_{95}$ $\downarrow$ & $\overline{\text{sIoU}}$ $\uparrow$ & $\overline{\text{PPV}}$ $\uparrow$ & $\overline{F_1}$ $\uparrow$ & AuPRC $\uparrow$ & FPR$_{95}$ $\downarrow$ & $\overline{\text{sIoU}}$ $\uparrow$ & $\overline{\text{PPV}}$ $\uparrow$ & $\overline{F_1}$ $\uparrow$\\
\midrule 
Maximum Softmax    & 28.0 & 72.1 & 15.5 & 15.3 & 5.4    
& 15.7 & 16.6 & 19.7 & 15.9 & 6.3 \\
MC Dropout         & 28.9 & 69.5 & 20.5 & 17.3 & 4.3    
& 4.9 & 50.3 & 5.5 & 5.8 & 1.1 \\ 
Ensemble           & 17.7 & 91.1 & 16.4 & \underline{20.8} & 3.4  
& 1.1 & 77.2 & 8.6 & 4.7 & 1.3 \\
PGN             & \underline{36.7} & \underline{61.4} & \underline{21.6} & 17.5 & \underline{6.2}  
& 16.5 & 19.7 & 19.5 & 14.9 & 7.4 \\
\midrule
ODIN               & 33.1 & 71.7 & 19.5 & 17.9 & 5.2  
& \underline{22.1} & 15.3 & \underline{21.6} & 18.5 & \underline{9.4} \\
Mahalanobis        & 20.0 & 87.0 & 14.8 & 10.2 & 2.7  
& 20.9 & \underline{13.1} & 13.5 & \underline{21.8} & 4.7 \\
\midrule
ours          & \textbf{81.0} & \textbf{9.9} & \textbf{53.8} & \textbf{26.7} & \textbf{13.6}
& \textbf{81.6} & \textbf{2.0} & \textbf{39.7} & \textbf{34.0} & \textbf{32.9} \\
\bottomrule
\end{tabular} }
\label{tab:ood_ao}
\end{table*}
%
%
%
%
\section{\uppercase{Conclusion}}\label{sec:conc}
In this work, we presented a multi-scale OOD segmentation method that exploits the confidence information of a foreground-background segmentation model. 
In comparison to supervised semantic segmentation models using a closed set of predefined classes, this independence from class prediction makes it reasonable to apply these models to the detection of unknown objects. 
We considered the per pixel confidence score which is close to 1 for a pixel in a foreground object and aggregated these confidence values for the different sized patches to identify objects of various sizes in a single image. 
To this end, we used different approaches for the multi-scale procedure, i.e., how the image patches of different sizes can be constructed and combined. 
Furthermore, we have observed that uncertainty information extracted from the softmax output of a DNN for semantic segmentation supports our method if the input images resemble the training images. 
Note, our approach does not require any additional training or auxiliary data. 
Our experiments have shown the ability of our approach to segment OOD objects of different datasets and to outperform comparable baselines in the SegmentMeIfYouCan benchmark.


\bibliographystyle{apalike}
{\small
\bibliography{example}}

\begin{thebibliography}{}

\bibitem[Ackermann et~al., 2023]{Ackermann2023}
Ackermann, J., Sakaridis, C., and Yu, F. (2023).
\newblock Maskomaly: Zero-shot mask anomaly segmentation.
\newblock In {\em 34th British Machine Vision Conference 2023, {BMVC}}. BMVA.

\bibitem[Besnier et~al., 2021]{Besnier2021}
Besnier, V., Bursuc, A., Picard, D., and Briot, A. (2021).
\newblock Triggering failures: Out-of-distribution detection by learning from local adversarial attacks in semantic segmentation.
\newblock In {\em IEEE/CVF International Conference on Computer Vision (ICCV)}.

\bibitem[Biase et~al., 2021]{Biase2021}
Biase, G.~D., Blum, H., Siegwart, R.~Y., and Cadena, C. (2021).
\newblock Pixel-wise anomaly detection in complex driving scenes.
\newblock {\em IEEE/CVF Conference on Computer Vision and Pattern Recognition (CVPR)}, pages 16913--16922.

\bibitem[Blum et~al., 2019a]{Blum2019}
Blum, H., Sarlin, P.-E., Nieto, J., Siegwart, R., and Cadena, C. (2019a).
\newblock Fishyscapes: A benchmark for safe semantic segmentation in autonomous driving.
\newblock In {\em IEEE/CVF International Conference on Computer Vision (ICCV) Workshops}.

\bibitem[Blum et~al., 2019b]{Blum2019_1}
Blum, H., Sarlin, P.-E., Nieto, J.~I., Siegwart, R.~Y., and Cadena, C. (2019b).
\newblock The fishyscapes benchmark: Measuring blind spots in semantic segmentation.
\newblock {\em International Journal of Computer Vision}, pages 3119 -- 3135.

\bibitem[Caron et~al., 2021]{Caron2021}
Caron, M., Touvron, H., Misra, I., Jegou, H., Mairal, J., Bojanowski, P., and Joulin, A. (2021).
\newblock Emerging properties in self-supervised vision transformers.
\newblock In {\em 2021 IEEE/CVF International Conference on Computer Vision (ICCV)}, pages 9630--9640.

\bibitem[Chan et~al., 2021a]{Chan2021_1}
Chan, R., Lis, K., Uhlemeyer, S., Blum, H., Honari, S., Siegwart, R., Fua, P., Salzmann, M., and Rottmann, M. (2021a).
\newblock Segmentmeifyoucan: A benchmark for anomaly segmentation.
\newblock In {\em Conference on Neural Information Processing Systems Datasets and Benchmarks Track}.

\bibitem[Chan et~al., 2021b]{Chan2021}
Chan, R., Rottmann, M., and Gottschalk, H. (2021b).
\newblock Entropy maximization and meta classification for out-of-distribution detection in semantic segmentation.
\newblock In {\em IEEE/CVF International Conference on Computer Vision (ICCV)}.

\bibitem[Chen et~al., 2018]{Chen2018}
Chen, L.-C., Zhu, Y., Papandreou, G., Schroff, F., and Adam, H. (2018).
\newblock Encoder-decoder with atrous separable convolution for semantic image segmentation.
\newblock In {\em European Conference on Computer Vision (ECCV)}.

\bibitem[Cordts et~al., 2016]{Cordts2016}
Cordts, M., Omran, M., Ramos, S., Rehfeld, T., Enzweiler, M., Benenson, R., Franke, U., Roth, S., and Schiele, B. (2016).
\newblock The cityscapes dataset for semantic urban scene understanding.
\newblock In {\em 2016 IEEE Conference on Computer Vision and Pattern Recognition (CVPR)}, pages 3213--3223, Los Alamitos, CA, USA. IEEE Computer Society.

\bibitem[Delić et~al., 2024]{Delic2024}
Delić, A., Grcić, M., and Šegvić, S. (2024).
\newblock Outlier detection by ensembling uncertainty with negative objectness.

\bibitem[Deng et~al., 2009]{Deng2009}
Deng, J., Dong, W., Socher, R., Li, L.-J., Li, K., and Fei-Fei, L. (2009).
\newblock Imagenet: A large-scale hierarchical image database.
\newblock In {\em 2009 IEEE Conference on Computer Vision and Pattern Recognition}, pages 248--255.

\bibitem[Gal and Ghahramani, 2016]{Gal2016}
Gal, Y. and Ghahramani, Z. (2016).
\newblock Dropout as a bayesian approximation: Representing model uncertainty in deep learning.
\newblock In {\em Proceedings of the 33rd International Conference on International Conference on Machine Learning}, volume~48, pages 1050--1059.

\bibitem[Galesso et~al., 2023]{Galesso2023}
Galesso, S., Argus, M., and Brox, T. (2023).
\newblock Far away in the deep space: Dense nearest-neighbor-based out-of-distribution detection.
\newblock In {\em IEEE International Conference on Computer Vision (ICCV), Workshops}.

\bibitem[Gao et~al., 2023]{Gao2023}
Gao, Z., Yan, S., and He, X. (2023).
\newblock Atta: Anomaly-aware test-time adaptation for out-of-distribution detection in segmentation.

\bibitem[Grcic et~al., 2021]{Grcic2021}
Grcic, M., Bevandić, P., and Segvic, S. (2021).
\newblock Dense anomaly detection by robust learning on synthetic negative data.

\bibitem[Grcic et~al., 2022]{Grcic2022}
Grcic, M., Bevandić, P., and Segvic, S. (2022).
\newblock Densehybrid: Hybrid anomaly detection for dense open-set recognition.
\newblock In {\em European Conference on Computer Vision (ECCV)}, pages 500--517.

\bibitem[Grcic et~al., 2023]{Grcic2023}
Grcic, M., Šarić, J., and Šegvić, S. (2023).
\newblock On advantages of mask-level recognition for outlier-aware segmentation.
\newblock {\em IEEE/CVF Conference on Computer Vision and Pattern Recognition Workshops (CVPRW)}, pages 2937--2947.

\bibitem[Gudovskiy et~al., 2023]{Gudovskiy2023}
Gudovskiy, D., Okuno, T., and Nakata, Y. (2023).
\newblock Concurrent misclassification and out-of-distribution detection for semantic segmentation via energy-based normalizing flow.
\newblock In {\em Proceedings of the Thirty-Ninth Conference on Uncertainty in Artificial Intelligence}. JMLR.org.

\bibitem[He et~al., 2016]{He2016}
He, K., Zhang, X., Ren, S., and Sun, J. (2016).
\newblock Deep residual learning for image recognition.
\newblock In {\em IEEE Conference on Computer Vision and Pattern Recognition (CVPR)}, pages 770--778.

\bibitem[Hendrycks and Gimpel, 2016]{Hendrycks2016}
Hendrycks, D. and Gimpel, K. (2016).
\newblock A baseline for detecting misclassified and out-of-distribution examples in neural networks.

\bibitem[Hümmer et~al., 2023]{Hummer2023}
Hümmer, C., Schwonberg, M., Zhou, L., Cao, H., Knoll, A., and Gottschalk, H. (2023).
\newblock Vltseg: Simple transfer of clip-based vision-language representations for domain generalized semantic segmentation.

\bibitem[Lakshminarayanan et~al., 2017]{Lakshminarayanan2017}
Lakshminarayanan, B., Pritzel, A., and Blundell, C. (2017).
\newblock Simple and scalable predictive uncertainty estimation using deep ensembles.
\newblock In {\em Proceedings of the 31st International Conference on Neural Information Processing Systems}, pages 6405--6416. Curran Associates Inc.

\bibitem[Lee et~al., 2020]{Lee2020}
Lee, H., Kim, S.~T., Navab, N., and Ro, Y. (2020).
\newblock Efficient ensemble model generation for uncertainty estimation with bayesian approximation in segmentation.

\bibitem[Lee et~al., 2018]{Lee2018}
Lee, K., Lee, K., Lee, H., and Shin, J. (2018).
\newblock A simple unified framework for detecting out-of-distribution samples and adversarial attacks.
\newblock {\em Neural Information Processing Systems (NeurIPS)}.

\bibitem[Liang et~al., 2018]{Liang2018}
Liang, S., Li, Y., and Srikant, R. (2018).
\newblock Enhancing the reliability of out-of-distribution image detection in neural networks.
\newblock {\em International Conference on Learning (ICLR)}.

\bibitem[Lis et~al., 2020]{Lis2020}
Lis, K., Honari, S., Fua, P., and Salzmann, M. (2020).
\newblock Detecting road obstacles by erasing them.

\bibitem[Lis et~al., 2019]{Lis2019}
Lis, K., Nakka, K., Fua, P., and Salzmann, M. (2019).
\newblock Detecting the unexpected via image resynthesis.
\newblock In {\em IEEE/CVF International Conference on Computer Vision (ICCV)}.

\bibitem[Liu et~al., 2023]{Liu2023}
Liu, Y., Ding, C., Tian, Y., Pang, G., Belagiannis, V., Reid, I., and Carneiro, G. (2023).
\newblock Residual pattern learning for pixel-wise out-of-distribution detection in semantic segmentation.
\newblock In {\em Proceedings of the IEEE/CVF International Conference on Computer Vision (ICCV)}, pages 1151--1161.

\bibitem[Maag et~al., 2023]{Maag2023_dataset}
Maag, K., Chan, R., Uhlemeyer, S., Kowol, K., and Gottschalk, H. (2023).
\newblock Two video data sets for~tracking and~retrieval of~out of~distribution objects.
\newblock In {\em Computer Vision {\textendash} {ACCV} 2022}, pages 476--494. Springer Nature Switzerland.

\bibitem[Maag and Riedlinger, 2024]{Maag2024_grads}
Maag, K. and Riedlinger, T. (2024).
\newblock Pixel-wise gradient uncertainty for convolutional neural networks applied to out-of-distribution segmentation.
\newblock {\em International Joint Conference on Computer Vision, Imaging and Computer Graphics Theory and Applications (VISAPP)}, pages 112--122.

\bibitem[Maag and Rottmann, 2023]{Maag2023_domain}
Maag, K. and Rottmann, M. (2023).
\newblock False negative reduction in semantic segmentation under domain shift using depth estimation.
\newblock In {\em Proceedings of the 18th International Joint Conference on Computer Vision, Imaging and Computer Graphics Theory and Applications}. {SCITEPRESS} - Science and Technology Publications.

\bibitem[MacKay, 1992]{Mackay1992}
MacKay, D. J.~C. (1992).
\newblock A practical bayesian framework for backpropagation networks.
\newblock {\em Neural Computation}, 4(3):448--472.

\bibitem[Mukhoti and Gal, 2018]{Mukhoti2020}
Mukhoti, J. and Gal, Y. (2018).
\newblock Evaluating bayesian deep learning methods for semantic segmentation.

\bibitem[Nayal et~al., 2023]{Nayal2023}
Nayal, N., Yavuz, M., Henriques, J.~a.~F., and G\"uney, F. (2023).
\newblock Rba: Segmenting unknown regions rejected by all.
\newblock In {\em Proceedings of the IEEE/CVF International Conference on Computer Vision (ICCV)}, pages 711--722.

\bibitem[Pinggera et~al., 2016]{Pinggera2016}
Pinggera, P., Ramos, S., Gehrig, S., Franke, U., Rother, C., and Mester, R. (2016).
\newblock Lost and found: detecting small road hazards for self-driving vehicles.
\newblock In {\em IEEE/RSJ International Conference on Intelligent Robots and Systems (IROS)}.

\bibitem[Rai et~al., 2023]{Rai2023}
Rai, S.~N., Cermelli, F., Fontanel, D., Masone, C., and Caputo, B. (2023).
\newblock Unmasking anomalies in road-scene segmentation.
\newblock In {\em Proceedings of the IEEE/CVF International Conference on Computer Vision (ICCV)}, pages 4037--4046.

\bibitem[Sim\'{e}oni et~al., 2023]{Simeoni2023}
Sim\'{e}oni, O., Sekkat, C., Puy, G., Vobecky, A., Zablocki, E., and P\'{e}rez, P. (2023).
\newblock Unsupervised object localization: Observing the background to discover objects.
\newblock In {\em 2023 IEEE/CVF Conference on Computer Vision and Pattern Recognition (CVPR)}, pages 3176--3186.

\bibitem[Sodano et~al., 2024]{Sodano2024}
Sodano, M., Magistri, F., Nunes, L., Behley, J., and Stachniss, C. (2024).
\newblock { Open-World Semantic Segmentation Including Class Similarity }.
\newblock In {\em 2024 IEEE/CVF Conference on Computer Vision and Pattern Recognition (CVPR)}, pages 3184--3194. IEEE Computer Society.

\bibitem[Tian et~al., 2022]{Tian2022}
Tian, Y., Liu, Y., Pang, G., Liu, F., Chen, Y., and Carneiro, G. (2022).
\newblock Pixel-wise energy-biased abstention learning for anomaly segmentation on complex urban driving scenes.
\newblock In {\em European Conference on Computer Vision (ECCV)}.

\bibitem[Vojíř and Matas, 2023]{Vojir2023}
Vojíř, T. and Matas, J. (2023).
\newblock Image-consistent detection of road anomalies as unpredictable patches.
\newblock In {\em IEEE/CVF Winter Conference on Applications of Computer Vision (WACV)}, pages 5480--5489.

\bibitem[Vojíř et~al., 2021]{Vojir2021}
Vojíř, T., \v{S}ipka, T., Aljundi, R., Chumerin, N., Reino, D.~O., and Matas, J. (2021).
\newblock Road anomaly detection by partial image reconstruction with segmentation coupling.
\newblock In {\em IEEE/CVF International Conference on Computer Vision (ICCV)}, pages 15651--15660.

\bibitem[Vojíř et~al., 2024]{Vojir2024}
Vojíř, T., Šochman, J., and Matas, J. (2024).
\newblock Pixood: Pixel-level out-of-distribution detection.

\bibitem[Wang et~al., 2023a]{Wang2023}
Wang, C.-Y., Bochkovskiy, A., and Liao, H.-Y.~M. (2023a).
\newblock Yolov7: Trainable bag-of-freebies sets new state-of-the-art for real-time object detectors.
\newblock In {\em 2023 IEEE/CVF Conference on Computer Vision and Pattern Recognition (CVPR)}, pages 7464--7475.

\bibitem[Wang et~al., 2023b]{Wang2023c}
Wang, X., Girdhar, R., Yu, S.~X., and Misra, I. (2023b).
\newblock Cut and learn for unsupervised object detection and instance segmentation.
\newblock In {\em IEEE/CVF Conference on Computer Vision and Pattern Recognition (CVPR)}, pages 3124--3134. IEEE.

\bibitem[Wortsman et~al., 2022]{Wortsman2022}
Wortsman, M., Ilharco, G., Gadre, S.~Y., Roelofs, R., Gontijo-Lopes, R., Morcos, A.~S., Namkoong, H., Farhadi, A., Carmon, Y., Kornblith, S., and Schmidt, L. (2022).
\newblock Model soups: averaging weights of multiple fine-tuned models improves accuracy without increasing inference time.
\newblock In Chaudhuri, K., Jegelka, S., Song, L., Szepesvari, C., Niu, G., and Sabato, S., editors, {\em Proceedings of the 39th International Conference on Machine Learning}, volume 162 of {\em Proceedings of Machine Learning Research}, pages 23965--23998. PMLR.

\bibitem[Xu et~al., 2023]{Xu2023}
Xu, J., Xiong, Z., and Bhattacharyya, S.~P. (2023).
\newblock Pidnet: A real-time semantic segmentation network inspired by pid controllers.
\newblock In {\em 2023 IEEE/CVF Conference on Computer Vision and Pattern Recognition (CVPR)}, pages 19529--19539. IEEE Computer Society.

\bibitem[Yan et~al., 2023]{Yan2023}
Yan, B., Jiang, Y., Wu, J., Wang, D., Luo, P., Yuan, Z., and Lu, H. (2023).
\newblock Universal instance perception as object discovery and retrieval.
\newblock In {\em 2023 IEEE/CVF Conference on Computer Vision and Pattern Recognition (CVPR)}, pages 15325--15336. IEEE Computer Society.

\bibitem[Zhang et~al., 2024]{Zhang2024}
Zhang, H., Li, F., Qi, L., Yang, M.-H., and Ahuja, N. (2024).
\newblock Csl: Class-agnostic structure-constrained learning for segmentation including the unseen.

\end{thebibliography}

\appendix
\section*{\uppercase{Appendix}}

%
%
\section{More Numerical Results}\label{sec:app_A}
The following experiments are conducted with FOUND as foreground-background model. 
In the main paper, we have uniformly combined the differently scaled confidence maps. In the following, we investigate whether a non-uniform weighting of the maps can further improve the OOD segmentation performance. 
We have tested several weightings using confidence maps with $N=1,16,64,256,1024$, see Table~\ref{tab:weightings}. 
The corresponding results are shown in Table~\ref{tab:ood_laf_found_non_uniform}. For a simple comparison, the best performing uniform combination is given as a baseline. 
We observe no or only minor improvements in individual metrics compared to the baseline.
\begin{table}[t]
\caption{Non-uniform weightings using confidence maps with $N=1,16,64,256,1024$.}
\centering
\scalebox{0.70}{
\begin{tabular}{l cccccc}
\toprule 
 & 1 & 16 & 64 & 256 & 1024 \\
\midrule
combination 1 & 0 & 0.25 & 0.35 & 0.2 & 0.2 \\
combination 2 & 0 & 0.3 & 0.4 & 0.2 & 0.1 \\
combination 3 & 0.2 & 0.25 & 0.4 & 0.1 & 0.05 \\
combination 4 & 0.05 & 0.1 & 0.4 & 0.25 & 0.2 \\
combination 5 & 0 & 0.4 & 0.4 & 0.2 & 0 \\
\bottomrule
\end{tabular} }
\label{tab:weightings}
\end{table}
\begin{table}[t]
\caption{OOD segmentation results on the LostAndFound dataset using different non-uniform combinations of the confidence maps.}
\centering
\scalebox{0.70}{
\begin{tabular}{l cc ccc}
\toprule 
 & \multicolumn{5}{c}{LostAndFound test-NoKnown} \\
\cmidrule(r){2-6} 
& AuPRC $\uparrow$ & FPR$_{95}$ $\downarrow$ & $\overline{\text{sIoU}}$ $\uparrow$ & $\overline{\text{PPV}}$ $\uparrow$ & $\overline{F_1}$ $\uparrow$ \\
\midrule
16+64+256+1024 & \textbf{56.5} & 10.6 & \textbf{32.2} & 22.5 & 19.5 \\
\midrule
combination 1 & 56.3 & 11.0 & 30.6 & 24.2 & 20.1 \\
combination 2 & 55.7 & 11.6 & 28.2 & \textbf{26.0} & \textbf{20.2} \\
combination 3 & 55.0 & 12.9 & 24.9 & 23.2 & 18.0 \\
combination 4 & 55.0 & \textbf{10.4} & 31.5 & 22.3 & 18.7 \\
combination 5 & 54.3 & 12.5 & 28.0 & 23.2 & 19.2  \\
\bottomrule
\end{tabular} }
\label{tab:ood_laf_found_non_uniform}
\end{table}
\begin{table}[!t]
\caption{OOD segmentation results on the LostAndFound dataset using patches of different sizes in one image.}
\centering
\scalebox{0.70}{
\begin{tabular}{l cc ccc}
\toprule 
 & \multicolumn{5}{c}{LostAndFound test-NoKnown} \\
\cmidrule(r){2-6} 
& AuPRC $\uparrow$ & FPR$_{95}$ $\downarrow$ & $\overline{\text{sIoU}}$ $\uparrow$ & $\overline{\text{PPV}}$ $\uparrow$ & $\overline{F_1}$ $\uparrow$ \\
\midrule
16+64+256+1024 & \textbf{56.5} & \textbf{10.6} & \textbf{32.2} & \textbf{22.5} & \textbf{19.5} \\
\midrule
patch scheme a  & 28.8 & 37.5 & 15.6 & 20.5 &  7.3 \\
patch scheme b  & 37.3 & 27.6 & 19.5 & 20.4 & 11.5 \\
patch scheme c  & 31.3 & 24.0 & 23.1 & 16.2 &  8.8 \\
\bottomrule
\end{tabular} }
\label{tab:ood_laf_found_dyn}
\end{table}

\begin{figure*}[t]
    \center
    \includegraphics[width=0.32\textwidth]{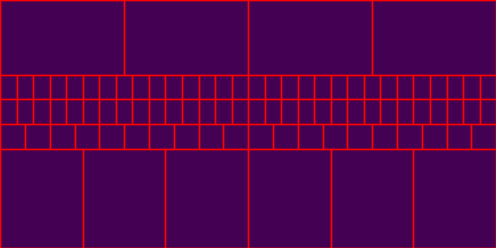}
    \includegraphics[width=0.32\textwidth]{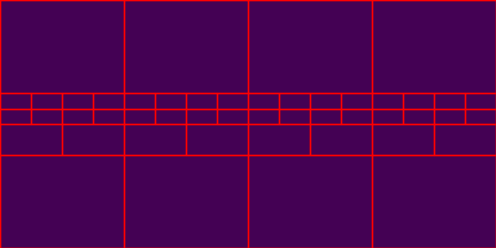}
    \includegraphics[width=0.32\textwidth]{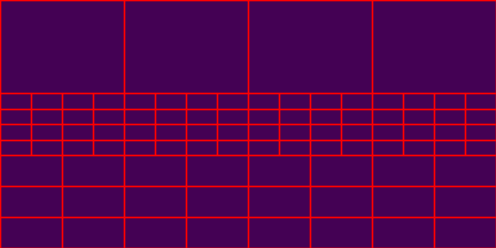} \\
    \vspace{0.5ex}
    \subfloat[][Patch scheme a]{\includegraphics[width=0.32\textwidth]{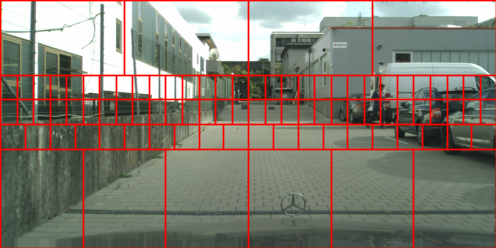} }
    \subfloat[][Patch scheme b]{\includegraphics[width=0.32\textwidth]{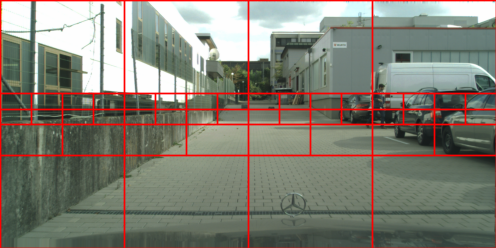} }
    \subfloat[][Patch scheme c]{\includegraphics[width=0.32\textwidth]{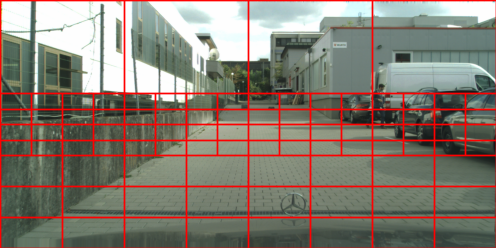} }
    \vspace{1ex}
    \caption{Three different patch schemes applied to the LostAndFound dataset.}
    \label{fig:patches}
\end{figure*}

Furthermore, we have conducted experiments where the confidence maps of the different scales are not combined, but patches of different sizes are applied in one image. In Figure~\ref{fig:patches}, three different patch schemes are illustrated. 
The idea is that the street scenes always resemble each other and the closer the objects are to the ego-car, the larger they are. 
The corresponding results are given in Table~\ref{tab:ood_laf_found_dyn}. 
We do not observe any improvement compared to the baseline, i.e., the uniform combination of different confidence maps. 

In Table~\ref{tab:ood_ao_found_entro}, the results for the RoadAnomaly21 and the RoadObstacle21 dataset for different number of patches $N$ as well as combinations of these confidence maps, and both in combination with the entropy heatmap obtained by the semantic segmentation network are shown. 
For both datasets, the entropy uncertainty heatmap cannot enhance the OOD segmentation performance. 
\begin{table*}[t]
\caption{OOD segmentation results on the RoadAnomaly21 and the RoadObstacle21 dataset with different number of patches $N$ as well as combinations of these confidence maps, and both in combination with the entropy heatmap obtained by the semantic segmentation network.}
\centering
\scalebox{0.70}{
\begin{tabular}{l cc ccc cc ccc}
\toprule 
\multicolumn{1}{c}{} & \multicolumn{5}{c}{RoadAnomaly21} & \multicolumn{5}{c}{RoadObstacle21} \\ 
\cmidrule(r){2-6} \cmidrule(r){7-11}
& AuPRC $\uparrow$ & FPR$_{95}$ $\downarrow$ & $\overline{\text{sIoU}}$ $\uparrow$ & $\overline{\text{PPV}}$ $\uparrow$ & $\overline{F_1}$ $\uparrow$ & AuPRC $\uparrow$ & FPR$_{95}$ $\downarrow$ & $\overline{\text{sIoU}}$ $\uparrow$ & $\overline{\text{PPV}}$ $\uparrow$ & $\overline{F_1}$ $\uparrow$\\
\midrule 
1    & 77.3 & \textbf{11.8} & 39.1 & \textbf{11.7} &  4.2      & 45.6 & 10.7 & 17.7 & 18.3 &  8.5  \\
4    & 73.0 & 16.0 & 44.8 & 10.4 &  4.0      & 54.7 &  7.8 & 22.5 & 21.5 & 13.2 \\
16   & 67.4 & 25.1 & \textbf{52.8} &  9.9 &  3.2      & 50.2 &  \textbf{5.2} & 24.0 & 20.5 & 13.1 \\
64   & 60.9 & 37.4 & 51.7 &  8.3 &  2.2      & 46.6 & 22.0 & 25.9 & 22.0 & 13.3 \\
256  & 44.3 & 60.2 & 38.2 &  9.4 &  1.0      & 35.0 & 44.0 & 27.4 & 15.8 &  7.9 \\
\midrule
1+4            & \textbf{76.0} & 12.7 & 43.7 & 11.3 & \textbf{4.4}       & 55.9 & 8.5 & 22.9 & 23.0 & 14.1 \\
1+4+16         & 74.7 & 14.5 & 47.8 &  9.8 & 3.9       & 57.9 & 7.2 & 26.4 & 20.1 & 14.7 \\
1+4+16+64      & 73.2 & 16.1 & 48.1 &  8.0 & 3.3       & \textbf{58.2} & 6.4 & 27.3 & 24.1 & \textbf{16.9} \\
16+64+256      & 62.7 & 28.0 & 47.4 &  7.3 & 1.9       & 51.8 & 5.5 & \textbf{28.5} & \textbf{24.5} & 16.4 \\
\bottomrule
\end{tabular} }
\label{tab:ood_ao_found_entro}
\end{table*}

\end{document}